\documentclass[sigconf]{acmart}

\acmConference[WWW]{ACM Web Conference}{May 2025}{Sydney, Australia}

\usepackage{lmodern}
\usepackage{enumitem}
\usepackage{hyperref}

\begin{document}

\title{Rethink Deep Learning with Invariance in Data Representation \\(Tutorial Proposal)}

\author{Shuren Qi}
\email{shurenqi@cuhk.edu.hk}
\affiliation{%
  \institution{The Chinese University of Hong Kong}
  \country{China}
}

\author{Fei Wang}
\email{few2001@med.cornell.edu}
\affiliation{%
  \institution{Cornell University}
  \country{United States}
}

\author{Tieyong Zeng}
\email{zeng@math.cuhk.edu.hk}
\affiliation{%
  \institution{The Chinese University of Hong Kong}
  \country{China}
}

\author{Fenglei Fan}
\email{flfan@math.cuhk.edu.hk}
\affiliation{%
  \institution{The Chinese University of Hong Kong}
  \country{China}
}

\renewcommand{\shortauthors}{S. Qi et al.}

\begin{abstract}
  Integrating \textbf{invariance} into data \textbf{representations} is a principled design in intelligent systems and web applications. Representations play a fundamental role, where systems and applications are both built on meaningful representations of digital inputs (rather than the raw data). In fact, the proper design/learning of such representations relies on priors w.r.t. the task of interest. Here, the concept of \emph{symmetry} from the \emph{Erlangen Program} may be the most fruitful prior — informally, a symmetry of a system is a transformation that leaves a certain property of the system invariant. Symmetry priors are ubiquitous, e.g., translation as a symmetry of the object classification, where object category is invariant under translation.

  The quest for invariance is as old as pattern recognition and data mining itself. Invariant design has been the cornerstone of various representations in \emph{the era before deep learning}, such as the SIFT. As we enter \emph{the early era of deep learning}, the invariance principle is largely ignored and replaced by a data-driven paradigm, such as the CNN. However, this neglect did not last long before they encountered bottlenecks regarding robustness, interpretability, efficiency, and so on. The invariance principle has returned in \emph{the era of rethinking deep learning}, forming a new field known as Geometric Deep Learning (GDL).

  In this tutorial, we will give a \textbf{historical perspective} of the invariance in data representations. More importantly, we will identify those research dilemmas, promising works, future directions, and web applications. 
  
\end{abstract}

\keywords{Pattern Recognition, Data Mining, Invariance, Symmetry, Representation, Tutorial}

\maketitle

\section{Topic and Relevance}

The topic of this tutorial is a historical review of the invariance in data representations. The scope of this tutorial covers 1) the invariance in the era before deep learning, on old-fashioned invariant designs from various hand-crafted representations; 2) the invariance in the early era of deep learning, on the slump of the invariance principle and the success of the data-driven paradigm; 3) the invariance in the era of rethinking deep learning, on the revival of the invariance principle and the emergence of geometric deep learning as a way to bridge the research gap. For the depth within each era, the research dilemmas, promising works, future directions, and web applications will be sorted out. \emph{More details are expanded in \textbf{Section 2}}.

The presenters are qualified for a high-quality introduction to the topic. We have extensive research experience and strong publication records in representation backbones and downstream applications of pattern recognition and data mining. \emph{More details are expanded in \textbf{Section 3}}.

This tutorial is timely, due to the general limitations of today's intelligent systems and their web applications with respect to being only data-driven. Also, the invariance perspective (technology focus) and the historical perspective (broad horizons) are rarely seen in the tutorial tracks of related conferences.

This tutorial is relevant to the Web Conference. From a technological perspective, representations play a fundamental role in intelligent systems and their wide range of downstream web applications. From a practical perspective, the currently popular data-driven paradigm has led to bottlenecks in intelligent systems and their web applications, regarding robustness, interpretability, efficiency, and so on. Understanding invariance in data representations is helpful in facilitating better web applications.

\section{Content}

Over the past decade, deep learning representations, e.g., convolutional neural networks (CNN) and transformer, have led to breakthrough results in numerous artificial intelligence (AI) tasks, e.g., processing human perceptual information, playing board games, and generating realistic media. Without exception, these successful programs are consistent with the principle of empirical risk minimization and rarely involve other realistic factors. More recently, their applications are expanding to more real-world scenarios, e.g., medical diagnostics, self-driving cars, online services, and web platforms. In such scenarios, the robustness, interpretability, and efficiency of AI systems are crucial: 1) robustness means the performance of system is stable for intra-class variations on the input; 2) interpretability means the behavior of system can be understood or predicted by humans; 3) efficiency means the real-time availability and energy cost during human-computer interaction.

Integrating invariant structures into representations is a principled design towards robust, interpretable, and efficient AI systems \cite{startref9}. Specifically, representations play a fundamental role, where the system is generally built on meaningful representations of digital inputs (rather than the raw data) \cite{startref10}. Note that the proper design/learning of such representations in fact relies on priors w.r.t. the task of interest. Here, the concept of symmetry from the \emph{Erlangen Program} \cite{startref11} may be the most fruitful prior — informally, a symmetry of a system is a transformation that leaves a certain property of system invariant \cite{startref12}. Symmetry priors are ubiquitous, e.g., translation as a symmetry of the object classification where object category is invariant under translation \cite{startref13}.

We focus on historical perspectives on the invariance or symmetry in the development of data representations \cite{startref14}. 

\begin{itemize}
    \item{Going back to the era before deep learning, symmetry priors (e.g., invariance and equivariance) w.r.t. geometric transformations (e.g., translation, rotation, and scaling) have been recognized as main ideas in designing representations \cite{ startref16}. However, these hand-crafted representations are all fixed in design, relying on (under)-complete dictionaries, and therefore fail to provide sufficient discriminability at larger scales, e.g., ImageNet classification task \cite{startref18}.}
    \item{As we enter the early era of deep learning, a cascade of learnable nonlinear transformations achieves over-complete representations of strong discriminative power for larger-scale pattern recognition and data mining tasks. As a textbook view now, representations should be learned not designed \cite{startref2}. Therefore, early learning representations are equipped with very few symmetry priors, typically just translation equivariance. Hence, these representations lack robustness, interpretability, and efficiency guarantees \cite{startref25}, e.g., the presence and understanding of adversarial perturbations \cite{startref20}. Note that the compatibility between invariance and discriminability has emerged as a tricky problem when moving towards real-world AI \cite{startref21}.}
    \item{Now in the era of rethinking deep learning, the invariance principle has returned, forming a new field known as Geometric Deep Learning (GDL) -- endowing a pattern recognition and data mining system with the basic symmetry structure of the physical world, and harmonizing knowledge-driven invariant representations and data-driven deep representations \cite{startref9}. The GDL research extends the scope of previous invariant theories from geometric, algebraic, and group, while showing the potential for uniformly improving the robustness, interpretability, and efficiency of the deep representation techniques \cite{startref4}.}
\end{itemize}

Despite the above GDL ideals, the current community of invariance faces the following research challenges:

\begin{itemize}
	\item{At the theoretical level, classical invariance is based on certain global assumptions. As for more informative local and hierarchical invariants in pattern recognition and data mining (i.e., going partial and deep), there is a challenge on corresponding theoretical expansion.}
	\item{At the practical level, classical invariance is often used in certain low-level pattern recognition and data mining tasks. As for higher-level tasks with symmetry prior, there is a challenge on corresponding practical expansion.}
\end{itemize}

We can respond to the above challenges with a long-term research. One idea is to extend the successful structure of modern deep learning to knowledge approach, exploring the discriminative potential of hand-crafted representations. Another idea is to embed the experience of invariant theory into modern learning representations, replacing the black-box nature of typical learning with controllability. Such theoretical efforts at the representation level are expected to have a wide range of web-related applications.
\begin{itemize}
	\item{Pattern recognition and data mining on graphs.} A uniqueness of geometric deep learning is the applicability to data types, covering typical grids and surfaces, as well as generalized sets and graphs. A good representation on sets or graphs implies, in fact, a full exploitation of their complex geometric priors. Due to the nodes and connectivity properties that are natural to the web, a range of web problems relies on invariant representations of sets or graphs.
        \item{Recommender systems and social networks.} Recommender systems are typically implemented by data mining on data of user-user graph (social networks), user-item graph (interactions), and item-item graph. Invariance is an important prior to achieve compact representations of graph data.
        \item{Cybersecurity and information forensics.} The automated generation and online distribution of misinformation is a real threat. Unlike typical tasks, cybersecurity tasks have a distinctly adversarial nature: the adversary actively creates obstacles to forensics, i.e., escape attacks. Invariance is an important prior to achieve robust and interpretable representations against escape attacks.
        \item{Efficient web services.} As deep learning models are used more on the server side, high inference costs have become a significant barrier for network service providers. Invariance is an important prior to avoid over-parameterization in the representation model on the server side.
        \item{Non-learning end-side deployments.} Even on the end side where computational resources are limited, sometimes expensive deep models are unavoidable due to the domain adaptive capabilities of end-to-end learning.  Invariance is an important prior to achieve non-learned but task-adaptive representations, by compressing discriminative-irrelevant feature spaces.
     \item{Physical consistency enhancement for large generative models.} Automatic content generation is currently a popular web application due to the fact that large models are usually only suitable to be placed on the server side. It has been known that large generative models are inefficient and unprotected in fitting physical structures. Invariance is an important prior to constrain large model representations to conform to physical and geometric patterns.

\end{itemize}

\section{Organizers}
\subsection*{Shuren Qi}
Dr. Shuren Qi is currently a Postdoctoral Fellow with Department of Mathematics, The Chinese University of Hong Kong. His research focuses on robust and explainable representations in Geometric Deep Learning, with applications in Trustworthy AI and Science AI. He has authored 14 papers in top-tier venues such as ACM Computing Surveys and IEEE Transactions on Pattern Analysis and Machine Intelligence. His works offer new designs of invariant representations — from global to local and hierarchical assumptions. The following research records by Shuren Qi are directly relevant to the tutorial topic:

\begin{itemize}[itemsep=-0pt]

\item S. Qi, Y. Zhang, C. Wang, J. Zhou, X. Cao. A survey of orthogonal moments for image representation: Theory, implementation, and evaluation. ACM Computing Surveys 55 (1), 1-35, 2023.

\item S. Qi, Y. Zhang, C. Wang, J. Zhou, X. Cao. A principled design of image representation: Towards forensic tasks. IEEE Transactions on Pattern Analysis and Machine Intelligence 45 (5), 
5337 - 5354, 2022.

\item S. Qi, Y. Zhang, C. Wang, T. Xiang, X. Cao, Y. Xiang. Representing noisy image without denoising. IEEE Transactions on Pattern Analysis and Machine Intelligence, 2024, in press.

\item S. Qi, Y. Zhang, C. Wang, Z. Xia, X. Cao, J Weng, F. Fan. Hierarchical invariance for robust and interpretable vision tasks at larger scales, arXiv preprint (under review), 2024.

\item C. Wang, S. Qi*, Z. Huang, Y. Zhang, R. Lan, X. Cao, F. Fan. Spatial-frequency discriminability for revealing adversarial perturbations. IEEE Transactions on Circuits and Systems for Video Technology, 2024, in press.

\end{itemize}

Presentations
\begin{itemize}[itemsep=-0pt]
\setcounter{enumi}{19}
\item The Invariant Representation Towards Trustworthy Artificial Intelligence, invited talks at The Chinese University of Hong Kong and Harbin Institute of Technology.

\end{itemize}

\subsection*{Fei Wang}

Dr. Fei Wang is currently an Associate Professor of Health Informatics in Department of Population Health Sciences, Weill Cornell Medicine, Cornell University. His major research interest is data mining and its applications in health data science. He has published more than 250 papers in AI and medicine, which have received more than 21.3K citations (GoogleScholar: \href{https://scholar.google.com/citations?hl=en&user=FjCbjDYAAAAJ}{link}). His H-index is 70. His papers have won 8 best paper awards at top international conferences on data mining and medical informatics. His team won the championship of the NIPS/Kaggle Challenge on Classification of Clinically Actionable Genetic Mutations in 2017 and Parkinson's Progression Markers' Initiative data challenge organized by Michael J. Fox Foundation in 2016. Dr. Wang is the recipient of the NSF CAREER Award in 2018, the inaugural research leadership award in IEEE International Conference on Health Informatics (ICHI) 2019. Dr. Wang is the chair of the Knowledge Discovery and Data Mining working group in American Medical Informatics Association (AMIA). 

The following research records by Dr. Fei Wang are directly relevant to the tutorial topic:

\begin{itemize}[itemsep=-0pt]

\item F. Wang, R. Kaushal, and D. Khullar. Should health care demand interpretable artificial intelligence or accept “black box” medicine? Annals of Internal Medicine 172, 59-60, 2020.

\item F. Wang, L. P. Casalino, and D. Khullar. Deep learning in medicine-promise, progress, and challenges. JAMA internal medicine 179(3), 293-294, 2019.

\item C. Zang, F. Wang. MoFlow: An invertible flow model for generating molecular graphs. KDD 2020.

\item C. Zang, F. Wang. Differential deep learning on graphs and its applications. KDD 2020.

\item M. Sun, S. Zhao, C. Gilvary, O. Elemento, J. Zhou and F. Wang. Graph convolutional networks for computational drug development and discovery. Briefings in bioinformatics, 21(3), 919-935. 2020. 

\end{itemize}

Tutorials
\begin{itemize}[itemsep=-0pt]
\setcounter{enumi}{19}

\item AI in Precision Medicine: Towards Knowledge Empowered Intelligence over ``Small” Data. Tutorial at 34th AAAI. 2020. (4 hours)

\item Recent Advances in Graph Analytics and Its Applications in Healthcare. Tutorial at the 26th KDD. 2020 (4 hours, with Peng Cui, Jian Pei, Yangqiu Song, Chengxi Zang)

\item Differential Deep Learning on Graphs and its Applications. Tutorial at 34th AAAI. 2020. (4 hours, with Chengxi Zang)

\item Learning From Networks: Algorithms, Theory, \& Applications. Tutorial at the 25th KDD. 2019 (7 hours, with Xiao Huang, Peng Cui, Yuxiao Dong, Jundong Li, Huan Liu, Jian Pei, Le Song, Jie Tang, Hongxia Yang, Wenwu Zhu)

\item Data Analytics with Electronic Health Records. Tutorial at 29th AAAI. 2015. (4 hours)

\item Data Analytics in Healthcare: Problems, Solutions and Challenges. Tutorial at 23rd CIKM, 2014. (3 hours)

\item Feature Engineering for Health Informatics. Tutorial at The 18th PAKDD. Tainan, Taiwan. 2014 (3 hours)
\end{itemize}

Dr. Fei Wang has served as senior program committee member/area chair of conferences including AAAI, IJCAI, KDD, CIKM, ICDM, SDM. He also reviews major medical and interdisciplinary journals including Nature Medicine, Nature Communications, Annals of Internal Medicine, etc.

\subsection*{Tieyong Zeng}

Dr. Tieyong Zeng is currently a Professor at the Department of Mathematics, The Chinese University of Hong Kong. Together with colleagues, he has founded the Center for Mathematical Artificial Intelligence (CMAI) since 2020 and served as the director of CMAI. His research interests include image processing, optimization, artificial intelligence, scientific computing, computer vision, machine learning, and inverse problems. He has published around 100 papers in the prestigious journals such as IEEE Transactions on Pattern Analysis and Machine Intelligence and International Journal of Computer Vision. He is laureate of the 2021 Hong Kong Mathematical Society (HKMS) Young Scholars Award.

The following research records by Dr. Tieyong Zeng are directly relevant to the tutorial topic:

\begin{itemize}[itemsep=-0pt]

\item J. Liu, M. Yan, and T. Zeng. Surface-aware Blind Image Deblurring. IEEE Transactions on Pattern Analysis and Machine Intelligence, 43(3), 1041-1055, 2021.

\item J. Liu, W. Liu, J. Sun, T. Zeng. Rank-one prior: Real-time scene recovery, IEEE Transactions on Pattern Analysis and Machine Intelligence, 45(7), 8845-8860, 2022.

\end{itemize}

\subsection*{Fenglei Fan}
Dr. Fenglei Fan is currently a Research Assistant Professor with Department of Mathematics, The Chinese University of Hong Kong. His primary research interests lie in NeuroAI and data mining. He has authored 26 papers in flagship AI and medical imaging journals. He was the recipient of the IBM AI Horizon Scholarship. He was also selected as the award recipient for the 2021 International Neural Network Society Doctoral Dissertation Award. His primarily-authored paper was selected as one of few 2024 CVPR Best Paper Award Candidates (26 out of 1W+ submissions) and won the IEEE Nuclear and Plasma Society Best Paper Award.

The following research records by Dr. Fenglei Fan are directly relevant to the tutorial topic:

\begin{itemize}[itemsep=-0pt]

\item F. Fan, J. Xiong, M. Li, G. Wang. On interpretability of artificial neural networks: A survey. IEEE Transactions on Radiation and Plasma Medical Sciences, 5(6), 741-760, 2021.

\item Z. Dong, F. Fan*, W. Liao, J. Yan. Grounding and enhancing grid-based models for neural fields, CVPR, 2024, in press (This paper gets full-graded review, CVPR 2024 Best Paper Award Candidate).

\item F. Fan, R. Lai, G. Wang. Quasi-equivalency of width and depth of neural networks. Journal of Machine Learning Research, 2023, in press.

\item F. Fan, M. Li, R. Lai, F. Wang, G. Wang. On Expressivity and Trainability of Quadratic Networks. IEEE Transactions on Neural Networks and Learning Systems, 2023.

\end{itemize}

Tutorial
\begin{itemize}[itemsep=-0pt]
\setcounter{enumi}{19}
\item Introducing Neuronal Diversity into Deep Learning. Tutorial at 37th AAAI. 2023. (1 hour and 45 minutes)
\end{itemize}

\section{Schedule}

\begin{itemize}
    \item Introduction (20 min)
    \begin{itemize}
        \item The bottlenecks of deep learning
        \item The potential of invariance
    \end{itemize}
    \item Preliminaries of invariance (20 min)
    \begin{itemize}
        \item Concepts of invariance and symmetry in physics and mathematics
        \item Cases of invariance and symmetry in pattern recognition and data mining tasks
        \item Formalizations of invariance and symmetry
    \end{itemize}
    \item Invariance in the era before deep learning (40 min)
    \begin{itemize}
        \item Invariance of global representations
        \item Invariance of local sparse representations
        \item Invariance of local dense representations
    \end{itemize}
    \item Invariance in the early era of deep learning (40 min)
    \begin{itemize}
        \item Hierarchical representations with data driven
        \item Invariance of hierarchical representations -- symmetry breaking
    \end{itemize}
    \item Invariance in the era of rethinking deep learning (40 min)
    \begin{itemize}
        \item Geometric deep learning as a way to bridge the gap
        \item The good, the bad, and the ugly
    \end{itemize}
    \item Conclusions and discussions (20 min)
    \begin{itemize}
        \item Review this tutorial
        \item Highlight research opportunities
    \end{itemize}
\end{itemize}

\section{Style and Audience}

This tutorial is lecture style. The intended audience for this tutorial mainly include researchers, graduate students, and industrial practitioners who are interested in exploring a new dimension of data representations.

The audience is expected to have the basic knowledge on deep learning, pattern recognition, and data mining. However, the tutorial will be presented at
college junior/senior level and should be comfortably followed
by academic researchers and industrial practitioners.

After this tutorial, the audience are expected to 1) have a comprehensive understanding of basic invariance concepts; 2) learn a history of invariance before and after the era of deep learning; 3) know the bottlenecks in the data-driven-only paradigm, the revival of invariance, and the geometric deep learning; and 4) explore novel research opportunities in this area, and master how to use or even design invariance representations for their tasks.

\section{Tutorial Materials}
This tutorial provides attendees with lecture slides, as well as a reading list of selected papers. We would like to state that such materials are free of any copyright issues.

\section{Video Teaser}
This tutorial has a video trailer, available at \href{https://www.dropbox.com/scl/fo/yj261r2tefsatajs83ju5/ACIcJmjCZXqhFhvu2pLvu5w?rlkey=noxe7toqy7konflqq9kbcy05r&st=q9241kft&dl=0}{link}.

\bibliographystyle{plainnat}
\bibliography{www}

\end{document}